\documentclass[sigconf,authorversion,nonacm]{aamas} 
\usepackage{balance} % for balancing columns on the final page

\usepackage{hyperref}
\usepackage{url}
\usepackage{amsmath}
\usepackage{amsthm}
\usepackage{booktabs}
\usepackage{algorithm}
\usepackage{algorithmic}
\usepackage{xcolor}
\usepackage{subfigure}
\usepackage{comment}
\usepackage{multirow}

\usepackage{mathtools,eqparbox}

\newcommand{\indices}[2]{{% \indices{<rows>}{<columns>}
  \begin{array}{@{}r@{}}
    \scriptstyle #2~\smash{\eqmakebox[ind]{$\scriptstyle\rightarrow$}} \\[-\jot]  
    \scriptstyle #1~\smash{\eqmakebox[ind]{$\scriptstyle\downarrow$}}
  \end{array}}}

\DeclarePairedDelimiterX{\infdivx}[2]{(}{)}{%
  #1\;\delimsize\|\;#2%
}

\setcopyright{ifaamas}
\acmConference[AAMAS '21]{Proc.\@ of the 20th International Conference on Autonomous Agents and Multiagent Systems (AAMAS 2021)}{May 3--7, 2021}{Online}{U.~Endriss, A.~Now\'{e}, F.~Dignum, A.~Lomuscio (eds.)}
\copyrightyear{2021}
\acmYear{2021}
\acmDOI{}
\acmPrice{}
\acmISBN{}

\title[Efficiently Guiding Imitation Learning Agents with Human Gaze]{Efficiently Guiding Imitation Learning Agents with Human Gaze}

\author{Akanksha Saran}
\affiliation{
  \department{Department of Computer Science}
  \institution{University of Texas at Austin}
  }
\email{asaran@cs.utexas.edu}

\author{Ruohan Zhang}
\affiliation{
  \department{Department of Computer Science}
  \institution{University of Texas at Austin}
  }
\email{zharu@utexas.edu}

\author{Elaine S. Short}
\affiliation{
  \department{Department of Computer Science}
  \institution{Tufts University}
  }
\email{elaine.short@tufts.edu}

\author{Scott Niekum}
\affiliation{
  \department{Department of Computer Science}
  \institution{University of Texas at Austin}
  }
\email{sniekum@cs.utexas.edu}

\begin{abstract}
Human gaze is known to be an intention-revealing signal in human demonstrations of tasks.  In this work,  we use gaze cues from human demonstrators to enhance the performance of agents trained via three popular imitation learning methods --- behavioral cloning (BC), behavioral cloning from observation (BCO), and Trajectory-ranked Reward EXtrapolation (T-REX).  Based on similarities between the attention of reinforcement learning agents and human gaze, we propose a novel approach for utilizing gaze data in a  computationally efficient manner,  as part of an auxiliary loss function, which guides a network to have higher activations in image regions where the human’s gaze fixated.  This work is a step towards augmenting any existing convolutional imitation learning agent's training with auxiliary gaze data. Our auxiliary coverage-based gaze loss (CGL) guides learning toward a better reward function or policy,  without adding any additional learnable parameters and without requiring gaze data at test time. We find that our proposed approach improves the performance by 95\% for BC, 343\% for BCO, and 390\% for T-REX, averaged over 20 different  Atari games. We also find that compared to a prior state-of-the-art imitation learning method assisted by human gaze (AGIL), our method achieves better performance, and is more efficient in terms of learning with fewer demonstrations. We further interpret trained CGL agents with a saliency map visualization method to explain their performance. At last, we show that CGL can help alleviate a well-known causal confusion problem in imitation learning.
\end{abstract}


\newcommand{\BibTeX}{\rm B\kern-.05em{\sc i\kern-.025em b}\kern-.08em\TeX}

\begin{document}

%%% The following commands remove the headers in your paper. For final 
%%% papers, these will be inserted during the pagination process.

\pagestyle{fancy}
\fancyhead{}

%%% The next command prints the information defined in the preamble.

\maketitle 

%%%%%%%%%%%%%%%%%%%%%%%%%%%%%%%%%%%%%%%%%%%%%%%%%%%%%%%%%%%%%%%%%%%%%%%%

\section{Introduction}
Learning agents can outperform humans at tasks such as Atari game playing when provided with well-defined goals or rewards using reinforcement learning (RL) \cite{sutton2018reinforcement,mnih2015human,silver2016mastering}. 
However, designing reward functions by hand can be difficult for complex tasks, even for experts. Imitation learning (IL) \cite{schaal1997learning,argall2009survey} is an alternative methodology which overcomes this difficulty by inferring an optimal policy from demonstrations. 
Additionally, imitation learning is a methodology which is intuitive and natural for novice end-users to train agents, similar to how humans teach other humans. %varied data sources to learn from
A challenge in training and utilizing IL agents in the real world is learning from few demonstrations to minimize the burden on end-users, while also sufficiently resolving ambiguity in user intentions and avoiding overfitting.  
Gaze, an additional informative modality from the demonstrator, can help extract more information out of the same number of demonstrations \cite{zhang2018agil}, in addition to information from state-action pairs.

\begin{figure}
\centering
\subfigure[Input image stack]{
\includegraphics[width=0.45\textwidth]{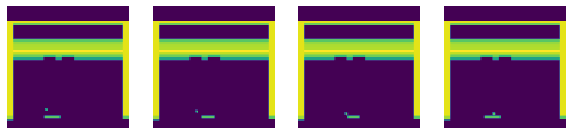}
}
\subfigure[Gaze heatmap]{
\includegraphics[width=0.11\textwidth]{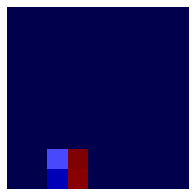}
}
\hspace{5mm}
\subfigure[Network activation without gaze loss]{
\includegraphics[width=0.11\textwidth]{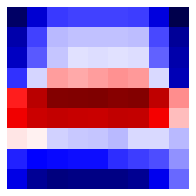}
 }
 \hspace{5mm}
 \subfigure[Network activation with gaze loss]{
\includegraphics[width=0.11\textwidth]{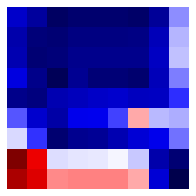}
 }
\caption{Our auxiliary gaze loss (CGL) guides a convolutional network to focus on parts of the state space which the human attends to (b). Examples of network activations for the BCO network on the Breakout game (a) are shown in (c) and (d). Activation map of the standard BCO algorithm not utilizing gaze and hence not required to attend to the area of human attention is shown in (c). Activation map of BCO with CGL incorporated as part of the training, which also attends to the area of human attention is shown in (d). 
}
\label{fig:activations}
\vspace{-3mm}
\end{figure}

Human attention in the form of eye gaze has been known to encode top-down attention versus bottom-up salience when performing goal-directed tasks \cite{land2009vision,hayhoe2005eye,rothkopf2007task,tatler2011eye}.
Gaze has been shown to improve the performance of imitation learning algorithms~\cite{zhang2019leveraging}, particularly for autonomous driving \cite{chen2019gaze,xia2019periphery} and Atari game playing \cite{zhang2018agil}. However, most prior approaches utilizing gaze for IL algorithms either use gaze heat maps as input to the agent's learning model in addition to the world state \cite{zhang2018agil,liu2019gaze}, or predict gaze heatmaps in conjunction with learning the policy \cite{xia2019periphery}. 
By contrast, we propose using an auxiliary gaze loss during training of imitation learning algorithms to improve the performance of existing methods without increasing model complexity, data requirements, or requiring test-time gaze.

Our methodology utilizes a demonstrator's gaze fixations on the image as part of a surrogate loss function (coverage-based gaze loss or CGL) during the training phase. Encoding priors in loss functions for label-free supervision of neural networks has been suggested by Stewart et al. \cite{stewart2017label}. Similarly, we use an auxiliary gaze loss to guide the learning of any agent using image-based state representations and convolutional layers as part of its model architecture.  Inspired by our experimental results highlighting the similarity of RL agent attention and human attention (Sec. \ref{sec:rl_attn}), we propose a coverage-based gaze loss (CGL). CGL guides a network to attend to the demonstrator's gaze locations and helps improve the performance of three IL methods on 20 Atari games. A critical advantage of our approach in contrast to several prior approaches utilizing gaze, is that gaze is not required at test time and instead used as a weak supervisory signal.

We evaluate our auxiliary gaze loss function on 20 Atari games with three different IL approaches. Our experiments show that CGL can improve performance for both inverse reinforcement learning (IRL) and behavioral cloning (BC) frameworks averaged across 20 games: 95\% for behavioral cloning (BC) \cite{zhang2018agil}, 343\% for behavioral cloning from observation (BCO) \cite{torabi2018behavioral}, and  390\% for T-REX \cite{brown2019extrapolating}, compared to not using any gaze information at all. Moreover, we show that to improve performance, human gaze is more informative than information already encoded in the visual state space in the form of motion of the visual scene. 

We also show that CGL outperforms two baseline methods that incorporate gaze information for imitation learning: (1) gaze-modulated dropout (GMD) \cite{chen2019gaze} and (2) attention guided imitation learning (AGIL) \cite{zhang2018agil}. 
Similar to our approach, GMD does not use additional learning parameters, whereas AGIL does.
We find that compared to AGIL, our auxiliary gaze loss is more efficient in incorporating gaze i.e. CGL improves performance more in low data regimes and does not require additional learnable parameters. We provide an analysis that shows that much of the performance improvement in AGIL comes from an increased number of model parameters and access to test-time gaze data, neither of which are required by CGL. 
We further perform analyses to show CGL indeed successfully guides learning agents to attend to important regions predicted by human gaze models through saliency map visualization~\cite{greydanus2018visualizing}. 
Finally, we also show experimental results that explain part of the gains with CGL can come from its ability to partially eliminate causal confusion.

\section{Related Work}

\subsection{Imitation Learning for Atari Games}
When learning from demonstrations, Atari game playing has been attempted with various imitation learning approaches. Behavioral cloning (BC) \cite{bain1999framework,ross2011reduction,daftry2016learning} is a class of imitation learning methods, where an agent learns a policy by using the demonstrated states and actions from the expert as input data for supervised learning. 
Behavioral cloning from observation (BCO) \cite{torabi2018behavioral} is a two-phase, iterative imitation learning technique -- first allowing the agent to acquire self-supervised experience in a task-independent pre-demonstration phase, which is then used to learn a model for a specific task policy only from state observations of expert demonstrations (without access to actions). The self-supervision produces an inverse dynamics model to infer actions, given state observations. This model is then used to infer expert actions from state-only demonstrations. The inferred actions along with state information are then used to perform imitation learning for the agent's policy. GAIL \cite{ho2016generative} is an adversarial imitation learning approach trained by alternating the learning updates between a generator policy network and a discriminator network distinguishing between the demonstrated and generated trajectories. It achieved state-of-the-art performance for low-dimensional domains. BCO shows comparable performance to GAIL on low-dimensional MuJoCo benchmarks \cite{todorov2012mujoco} with increased learning speed.

Behavioral cloning does not explicitly model the goals or intentions of the demonstrations which a succinct reward function attempts to capture in inverse reinforcement learning (IRL), another class of imitation learning approaches. Typically, such a succinct inferred reward function makes IRL have better generalization properties compared to behavioral cloning \cite{ross2011reduction}. The inferred reward function of the demonstrator can then be used by RL algorithms to learn the optimal policy. 
Most deep learning-based IRL methods either require access to demonstrated actions \cite{ibarz2018reward} or do not scale to high-dimensional tasks such as video games \cite{finn2016guided,fu2017learning,qureshi2018adversarial}. 
Tucker et al. \cite{tucker2018inverse} showed that their adversarial IRL method is difficult to train and fails at high-dimensional tasks of Atari game playing, even with extensive parameter tuning. Aytar et al. \cite{aytar2018playing} learn a reward function from observations for three Atari games. They guide the agent to exactly imitate the checkpoints from provided demonstrations, 
assuming access to high-quality demonstrations. A recent IRL method called T-REX \cite{brown2019extrapolating} is a reward learning from observation algorithm, that extrapolates beyond a set of ranked and potentially suboptimal demonstrations. T-REX outperforms other imitation learning methods such as BCO and GAIL, on Atari and MuJuCo benchmarks \cite{todorov2012mujoco} and also demonstrates the ability to extrapolate intentions of a suboptimal demonstrator. However, a performance gap in terms of the final scores achieved exists between reinforcement and imitation learning methods for Atari game playing. In our work, we propose to reduce this gap by incorporating an additional information modality in the form of human gaze for imitation learning.

\subsection{Utilizing Gaze for Learning}
Prior studies have shown that human fovea moves to the correct place at the right time to extract task-relevant information, making visual attention a feature selection mechanism for humans \cite{rothkopf2007task,hayhoe2005eye}. Human gaze information can be used in many ways to help AI agents learn a variety of tasks~\cite{zhang2020human}. Novice human learners can benefit from observing experts' gaze \cite{vine2012cheating} for learning complex surgical skills.  Yamani et al. \cite{yamani2017following} showed that viewing expert gaze videos can improve the hazard anticipation ability of novice drivers. 
Saran et al. \citep{saran2019understanding, saran2018human} showed the advantage of incorporating a human demonstrator's gaze for learning robotics manipulation tasks. Penkov et al. \cite{penkov2017physical} learn the mapping between abstract task plan symbols and their physical instances in the environment using eye gaze. 
Gaze has been exploited in prior imitation learning approaches for autonomous driving \cite{chen2019gaze,xia2019periphery} and Atari Games \cite{zhang2018agil}, but to the best of our knowledge, our work is the first attempt to incorporate gaze in a deep IRL algorithm (T-REX).% for Atari games.

A common method of incorporating human attention is to simply use the gaze map as an additional image-like input \cite{liu2019gaze} or predict the gaze heatmap and further use high-resolution parts of the image to improve learning \cite{zhang2018agil,xia2019periphery}. \citet{zhang2018agil} show improved learning on Atari games for imitation learning (AGIL) but use predicted heatmaps corresponding to demonstration states as part of the input to a BC network.  In comparison, CGL does not require any additional learnable parameters and can augment any existing imitation learning architecture. We compare the performance of CGL with BC against AGIL in our experiments.

Gaze-modulated dropout (GMD) was proposed by  \citet{chen2019gaze} to implicitly incorporate gaze into an IL framework for autonomous driving, instead of using gaze as an additional input. An estimated gaze distribution is used to modulate the dropout probability of units at different spatial locations in the first two convolutional layers. 
While GMD requires gaze both at train and test time, our auxiliary loss only requires gaze data at train time. Both GMD \cite{chen2019gaze} and our auxiliary gaze loss CGL do not increase the learnable parameters of an agent's network, and hence we use it as a baseline for comparison.  
\section{Approach}
To enable existing IL algorithms to take advantage of human gaze signals accompanying demonstrations, we first try to understand the correlation between human gaze and well-performing RL agents for six Atari games (Sec. \ref{sec:rl_attn}). Based on our findings, which show a similarity between human gaze and RL attention, we formalize our auxiliary coverage-based gaze loss term to mimic the attention mechanism of human demonstrators (Sec. \ref{sec:loss}). This loss term can guide any convolutional network to attend towards features that human demonstrators attend to. Our approach does not increase the model complexity of existing algorithms in terms of the number of learnable parameters, and can be easily applied to the training of any neural network with convolutional layers. We then describe other baseline approaches incorporating human gaze information for IL, which we evaluate and compare against our proposed approach (Sec.~\ref{sec:baselines}).

\subsection{Human Gaze Coverage of RL Agent Attention}
\label{sec:rl_attn}
\begin{figure}
\centering
\subfigure[\footnotesize Game State]{
\includegraphics[width=0.12\textwidth]{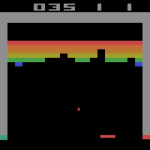}
}
\subfigure[\footnotesize RL Attention]{
\includegraphics[width=0.12\textwidth]{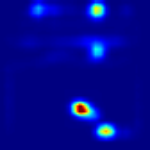}
}
\subfigure[\footnotesize Human Gaze]{
\includegraphics[width=0.12\textwidth]{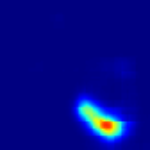}
 }
 
\caption{A trained PPO agent's attention map (b) and corresponding human gaze map (c) for the same input image (a) for the Breakout Atari game. The RL attention is directed towards more regions than the human gaze.}
\label{fig:rl_attn}
\vspace{-2mm}
\end{figure}

Prior studies in the cognitive science literature have established the concept of selective attention for decision making in primates \citep{petrosino2013selective}. While choosing an action among the available set of actions, animals select a subset of information by directing their sensory organs towards specific stimuli (overt attention) and focus on specific parts of the stimulus internally (covert attention) to act upon. %Overt and covert attention. 
Human gaze data only reveals \textit{overt} attention which is directly connected to a sensory organ. However, humans can still pay \textit{covert} attention to entities in the working memory \citep{posner1980orienting}. In other words, being attended by the human gaze model is a sufficient (but not necessary) condition for the features to be important. An example of this is shown in Fig. \ref{fig:rl_attn} where a high-performing RL agent on the game Breakout attends to more features of the state space in addition to what the human gaze attends to (human gaze predicted by a reliable generative human gaze model \cite{zhang2018agil}, RL attention calculated by~\cite{greydanus2018visualizing}). Hence for our analysis comparing the attention of RL agents and humans, we define a \textit{coverage} metric (Equation \ref{eq:coverage}) that penalizes the RL agent only if it fails to pay attention to human attended regions, or equivalently, a metric that is sensitive to false negatives if we treat human attention as the ground truth. KL divergence is an ideal candidate in this case \citep{bylinskii2018different}. 
Let P denote the human attention map and Q denote the RL attention map. The \textit{coverage} metric is computed as follows:
\begin{equation}
    KL(P||Q) = \sum_i \sum_j P(i,j)\log\bigg( \frac{P(i,j)+\epsilon }{Q(i,j)+\epsilon}\bigg)
    \label{eq:coverage}
\end{equation}

\noindent where $\epsilon$ is a small regularization constant chosen to be $2.2204E^{-16}$ following convention \citep{bylinskii2018different}. A lower value of the KL-divergence based \textit{coverage} metric would signify a stronger correlation between human gaze and RL attention.

We compute this \textit{coverage} metric between the RL attention map and the corresponding human gaze map for 100 images per game. The overall metric reported for a game is averaged over all 100 images. The images are uniformly sampled from a policy rollout of the learned RL agent trained via proximal policy optimization (PPO). To examine if there is a similarity between the attention of the agent and human, we compare human attention maps from one game with RL attention maps of the same game and the other five games. We hypothesize that the \textit{coverage} metric would have a lower value for the gaze maps and RL attention maps from the same game versus other games.

\begin{table*}
    \centering
    \caption{Comparison between PPO network attention and human gaze attention using KL divergence. The values represent the average KL divergence between gaze heat maps and RL attention heat maps for 100 uniformly sampled images from a policy rollout of the PPO agent. 
    }
    \begin{tabular}{|c|c|c|c|c|c|c|}
    \hline
    $\indices{\text{Human Attention}}{\text{RL Attention}}$ & asterix & breakout & centipede & ms\_pacman & phoenix & seaquest\\
    \hline
        asterix & \textbf{1.72}	&6.97	&4.64	&3.13&	14.61&	6.34\\
        \hline
        breakout & 5.26&	\textbf{2.09}&	4.94&	3.80&	11.13&	5.57\\
        \hline
        centipede & 4.43&	6.40&	\textbf{1.86}&	3.34&	9.86&	5.71\\
        \hline
        ms\_pacman & 4.53&	6.46&	5.49&	\textbf{1.78}&	13.34&	5.92\\
        \hline
        phoenix & 4.29&	10.75&	5.19&	3.59&	\textbf{3.55}&	6.49\\
        \hline
        seaquest & 5.07&	7.70&	5.97&	3.51&	14.10&	\textbf{3.03}\\
        \hline
    \end{tabular}
    \label{tab:compare}
\end{table*}

We find that the lowest average \textit{coverage} metric scores are obtained between RL attention heatmaps and gaze heatmaps for the same game (diagonal of Table \ref{tab:compare}). Comparison of human gaze heatmaps for one game and RL attention heatmaps of all other game agents is equivalent to comparison of gaze heatmaps with randomly generated attention maps (without knowledge of the game state with which gaze heatmap is computed). This one-to-one comparison for each of the 100 time steps along a trajectory implies the human gaze and RL attention for the same task are the most similar. The analysis here informs our auxiliary loss function to guide the attention of imitation learning algorithms to have \textit{coverage} over human gaze.

\subsection{Coverage-based Gaze Loss}
\label{sec:loss}
Based on the coverage metric (Equation \ref{eq:coverage}), we propose adding an auxiliary loss term in addition to the existing loss function for a network, modifying the training procedure of any IL algorithm. Our loss term will have a higher penalty if the network does not attend to parts of the image that the human demonstrator focused on, but will have no penalty for activations where the demonstrator did not pay attention. We refer to the proposed loss function as a coverage-based gaze loss (CGL). 

CGL operates on the human gaze heatmap and the output of the last convolutional layer. For consistency in comparison with baselines, gaze heatmaps are generated using convolution-deconvolution networks trained on real human gaze data \citep{zhang2018agil}. Activation feature maps from the last convolutional layer \citep{selvaraju2017grad} of image classification CNNs are shown to have the best compromise between high-level semantics and detailed spatial information. Given a 3D feature map of size $h\times w \times c$ from a convolutional layer, it is collapsed to a feature map $f$ of size $h\times w$ using a $1\times1$ convolutional filter.
Equation \ref{eq:2} shows the normalization of this feature map $f$ using a softmax operator to values between 0 and 1. Given a normalized 2D gaze heatmap $g$ of size $h\times w$, CGL is computed as:
% $$

\begin{equation}
CGL(g,f^{'})  = \sum_{i \in (1,h)} \sum_{j \in (1,w)} g_{i,j} \bigg{[} \log\frac{g_{i,j}+\epsilon}{f_{i,j}^{'}+\epsilon} \bigg{]} 
\label{eq:1}
\end{equation}

where
\begin{equation}
f_{i,j}^{'} = \frac{\exp^{f_{i,j}}}{\sum_{k=0}^{k=h-1}\sum_{j=0}^{j=w-1} \exp^{f_{k,l}}}
\label{eq:2}
\end{equation}

CGL adds a penalty if activations from none of the convolutional filters are high on areas where the demonstrator's gaze fixates during gameplay. 
Only regions of the gaze map which have a non-zero value contribute to the auxiliary loss, and other regions of the convolutional output which are not fixated on by the demonstrator do not affect the loss term. Hence, our loss term encourages \textit{coverage} of the demonstrator's attended regions. This is because unattended regions may also contain information necessary for decision-making \citep{petrosino2013selective}.

The magnitude of the penalty is computed using a smoothed ($\epsilon=2.2204E^{-16}$) KL divergence term between the normalized gaze map and the collapsed and normalized convolutional map, and is then weighted by the amount of gaze fixation an image region gets (Equation \ref{eq:1}).
Instead of forcing the filter weights to exactly match the demonstrator's gaze, CGL guides the network to focus on aspects of the state space which might be missed by the network, for example, areas of the image which are not feature-rich but are critical for decision-making (e.g., the ball in Fig.~\ref{fig:greydanus}(a)), eventually leading to better performance. A loss function which encourages a network to attend proportional to the human's gaze frequency instead, will be more restrictive. 

\subsubsection{Auxiliary Gaze Loss for BC}
For the behavioral cloning (BC) method, the gaze coverage loss is added as an auxiliary loss term in addition to the log likelihood action classification loss:
\begin{equation}
\mathcal{L}(\theta) = \sum_{i=1}^N \Bigg [-(1-\alpha)\log \pi_{\theta}(a_i | s_i) +  \alpha  \hspace{1mm}  CGL(g(s_i),c_3(s_i))\Bigg ] 
\end{equation}

The network architecture is similar to the one used in Zhang et al. \cite{zhang2018agil}, comprising of three convolutional layers and one fully-connected layer. It takes in a single game image as input and outputs a vector that gives the probability of each action. The gaze coverage loss is applied to the feature maps at the third convolutional layer. $g(s_i)$ is the gaze map of size $21\times21$, $c_3(s_i)$ is the collapsed and normalized feature map of size $21\times21$ (Equation \eqref{eq:2}) from the third convolutional layer.

\subsubsection{Auxiliary Gaze Loss for BCO}
For BCO \cite{torabi2018behavioral}, we incorporate CGL as part of learning the imitation policy after the agent learns an inverse-dynamics model of the environment. Similar to Torabi et al. \cite{torabi2018behavioral}, we use a neural network with three convolutional layers and one fully-connected layer using a stack of four consecutive frames as input. The output is the probability distribution over the discrete action space of the Atari domain. The network is learned using maximum likelihood estimation (MLE), finding the network parameters that best match the provided state-action pairs -- states $s_i$ obtained from a demonstrated trajectory $\tau_i$ and actions $\widetilde{a_i}$ recovered from the inverse dynamics model. The new loss function is a weighted combination of the standard cross-entropy loss for MLE and CGL applied to the output of the last convolutional layer as shown below.

\begin{equation}\label{eq:bco}
\mathcal{L}(\theta) = \sum_{i=1}^{N} \Bigg [ - (1-\alpha) \log \pi_{\theta}(\widetilde{a_i} | s_i)   +  \alpha \hspace{1mm} CGL(g(s_i), c_{3}(s_i)) \Bigg ] 
\end{equation}

Here, $\pi_{\theta}$ is the imitation policy network, $g(s_i)$ is the gaze map of size $84\times84$, $c_3(s_i)$ is the collapsed and normalized feature map (Equation \eqref{eq:2}) from the last convolutional layer ($7\times7$ size map upsampled to $84\times84$).

\subsubsection{Auxiliary Gaze Loss for T-REX}
T-REX \cite{brown2019extrapolating} is concerned with the problem of reward learning from observation, using rankings of demonstrations to efficiently infer a reward function. To the best of our knowledge, gaze has not been incorporated as part of a deep inverse reinforcement learning. 
Given a sequence of $m$ demonstrations ranked from worst to best,  $\tau_1, \dots ,\tau_m$, a parameterized reward network $\hat{r}_\theta$ is trained with a cross-entropy loss over a pair of trajectories ($\tau_i \prec \tau_j$), where $\tau_j$ is ranked higher than $\tau_i$. We add CGL to the reward network's loss, so the new loss function becomes:

\begin{equation} \label{eq:trex}
\begin{split}
&\mathcal{L}(\theta) =\\& (1-\alpha)\Bigg [-\sum_{\tau_i \prec \tau_j} \log \frac{\exp \sum_{s \in \tau_j}\hat{r}_\theta(s)}{\exp \sum_{s \in \tau_i}\hat{r}_\theta(s) + \exp \sum_{s \in \tau_j}\hat{r}_\theta(s)}\Bigg ]\\&+\alpha \Bigg [ \sum_{s \in \tau_i} CGL(\tau^{g}_{i}(s),c_{4}(s)) + \sum_{s \in \tau_j} CGL(\tau^{g}_{j}(s),c_{4}(s)) \Bigg ] 
\end{split}
\end{equation}

$\tau^{g}_{i}(s)$ represents the gaze map corresponding to the state $s$ from the trajectory snippet $\tau_i$ and $c_{4}(s)$ represents the collapsed and normalized version of the last  convolutional layer's output for the same state $s$. The loss function accumulates gaze over the entire trajectory snippet for both trajectories used as input to the network.

We use the default implementation of T-REX from Brown et al. \cite{brown2019extrapolating}. The reward network has four convolutional layers. 
The gaze loss is computed over the last convolutional layer output -- a spatial map of size $16\times7\times7$ (normalized, collapsed and upsampled to the size of the gaze heatmaps $84\times84$). At the end, a fully connected layer with 64 hidden units with a single scalar output is used to determine the ranking between a pair of demonstrations.  

Similar to the implementation of Brown et al. \cite{brown2019extrapolating}, the trajectories are first subsampled by maximizing over every 3rd and 4th frame, from which a stack of 4 consecutive frames with pixel values normalized between 0 and 1 is passed as input to the reward network. 
The snippets are ranked based on ground truth rewards or cumulative game scores of the trajectories they are sampled from. A PPO agent is then trained using the learned reward to obtain a policy for gameplay.

\subsection{Other Techniques to Incorporate Gaze}
\label{sec:baselines}
Here we describe two alternative methods incorporating human gaze for imitation learning, which we compare against.

\subsubsection{Gaze-modulated Dropout (GMD)}
\label{sec:baseline}
As a baseline for learning from human gaze, we implement GMD \cite{chen2019gaze} for the first two convolutional layers of the BCO policy network  \citet{torabi2018behavioral}. The BCO policy network does not originally use dropout layers. 
Gaze maps are generated using a convolution-deconvolution network \cite{zhang2018agil}, trained separately for each game on the Atari-HEAD dataset \cite{zhang2019atari}.  
The gaze prediction network uses as input a stack of 4 consecutive game frames, each of size $84\times84$. Details of the network architecture are similar to  \citet{zhang2018agil}. We employ this network for gaze prediction, as it has been shown to work well for the Atari domain, instead of the Pix2Pix network \cite{isola2017image} used by \citet{chen2019gaze} for the autonomous driving domain.
The generated gaze map is then used as a mask for the additional dropout layer added after first two the convolutional layers as described by Chen et al. \cite{chen2019gaze}. Units of the convolutional layer near the estimated gaze location are assigned a lower dropout probability than units far from the estimated gaze location. This is similar to conventional dropout \cite{srivastava2014dropout}, but with non-uniform dropout probability for spatial units corresponding to different parts of the image space. 

\subsubsection{Attention Guided Imitation Learning (AGIL)} AGIL adds more parameters to a BC network to utilize gaze. 
The output of the gaze prediction network  
is used as input to an additional convolutional pathway in a modified version of standard behavioral cloning. AGIL consists of two channels of 3 convolutional layers. One channel takes as input a single image frame (game state) and another uses a masked image which is an element-wise product of the original image and predicted gaze saliency map. Finally, the outputs of the two channels are averaged to predict one of the 18 actions within ALE \cite{bellemare2013arcade}. We use the same hyperparameters provided by Zhang et al. \shortcite{zhang2018agil} for the implementation of AGIL. 

\section{Experiments and Results}
We use demonstrations from 20 games in ALE \cite{bellemare2013arcade} with varying dynamics and features. Demonstrations and corresponding human gaze data are from the publicly available Atari-HEAD dataset \citep{zhang2019atari}. We augmented three imitation learning algorithms with CGL --- BC, BCO, and T-REX.  %\footnote{https://github.com/asaran/IL-CGL}. 
These algorithms were implemented in the OpenAI Gym platform \cite{brockman2016openai}, which contains Atari 2600 video games with high-dimensional observation space (raw pixels). All reported results were game scores averaged over 30 different rollouts (episodes) of the learned policy, similar to the procedure followed by \citet{zhang2018agil}. We used the default settings from OpenAI baselines \cite{baselines} for parameters of ALE \cite{bellemare2013arcade}. All experiments are conducted on server clusters with NVIDIA 1080, 1080Ti, Titan V, or DGX GPUs.

For evaluation, we intend to show improvement in terms of game scores using CGL. We calculate the improvement factor over baseline in the following way: improvement = (new score - baseline score)~/ baseline score.  If both the baseline score and the new score are zero, improvement is zero. However, for some games baseline game scores are zeroes but new scores are non-zero. In such cases, the improvement will not be calculated. We report average improvement (including games in which improvements are negative) across 20 games. Details on the experiments and individual games scores can be found in the Appendix. 
Note that the improvement factors are \emph{underestimated}, due to the way we handle zero score games.

\subsection{CGL Improves BC}
BC+CGL outperforms basic BC on 19 out of 20 games with 15 minutes of demonstration data (Fig.~\ref{fig:bc_improv}). On average, the improvement is \textbf{95.1\%} (Table \ref{tab:bc_15}). With all 300 minutes of human gameplay data, BC+CGL outperforms BC on all 20 games with an average improvement of \textbf{86.2\%} (Table \ref{tab:bc_300}). The hyperparameter  $\alpha$ is tuned using a grid search from a set of 7 values --- 0.01, 0.05, 0.1, 0.3, 0.5, 0.7, 0.9. Batch size $N=50$ and Adadelta optimizer \cite{zeiler2012adadelta} is used for all models based on BC.

\subsubsection{Efficiency of CGL in terms of Learnable Parameters}
Previous methods (such as AGIL) incorporate human attention by introducing extra parameters to the model due to additional neural network modules added. To tease apart whether improvement in these approaches comes from increased parameters to standard behavioral cloning or from the gaze information itself, we perform the following experiment. We re-train the AGIL network, but instead of using gaze heatmaps, we pass the original image as input to the gaze pathway, referred to as the BC-2ch model. This helps us disambiguate if more parameters in the model help more versus the gaze information itself. As shown in Fig.~\ref{fig:bc_improv}, 
we find that the BC-2ch model does result in improved performance over BC, indicating that part of AGIL's improvement over BC is due to additional parameters. This hints at the fact that increasing model complexity alone without using any additional information as input proves beneficial. 

\subsubsection{CGL Provides Stronger Guidance than AGIL}
\begin{figure}
    \centering
    \includegraphics[width=0.45\textwidth]{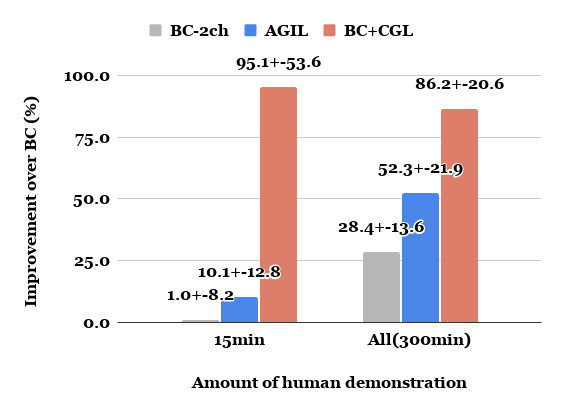}
    \caption{Average (across 20 games) percentage improvement over the BC baseline. Results are presented as as mean$\pm$standard error of the mean (N=20). Individual game scores of all agents can be found in Tables \ref{tab:bc_15} and \ref{tab:bc_300}.
    }
    \label{fig:bc_improv}
\end{figure}

In Fig.~\ref{fig:bc_improv}, we also show that on average, CGL outperforms the previously best method to incorporate gaze (AGIL~\cite{zhang2018agil}) by a large margin. This study suggests that CGL performs significantly better than AGIL with fewer model parameters, and the advantage is even more evident with a limited amount of human demonstration data (95.1\% improvement over BC versus 10.1\%). The sample efficiency of CGL is critical as it can be beneficial for applying this method to challenging imitation learning problems, where collecting demonstrations is cumbersome, expensive and time-critical. 

\subsection{CGL Improves BCO}
\begin{figure}
    \centering
    \includegraphics[width=0.45\textwidth]{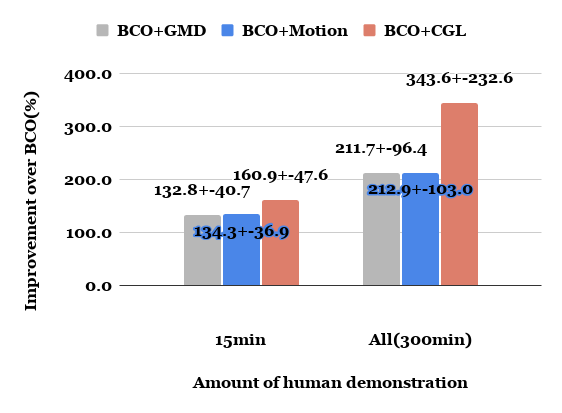}
    \caption{Average (across 20 games) improvement over the BCO baseline. Results are presented as mean$\pm$standard error of the mean (N=20). Individual game scores of all agents can be found in Tables \ref{tab:bco_15} and \ref{tab:bco_300}.
    } 
    \label{fig:bco_improv}
\end{figure}

BCO+CGL outperforms basic BCO on 14 out of 20 games with 15 minutes of human demonstration data (Fig.~\ref{fig:bco_improv}). On average, the improvement is \textbf{160.9\%}. The hyperparameter  $\alpha$ is tuned using a grid search from a set of 9 values --- 0.001, 0.005, 0.01, 0.05, 0.1, 0.3, 0.5, 0.7, 0.9. The Adam optimizer \cite{kingma2014adam} is used to solve for the network parameters with a batch size $N=32$. There are six games for which BCO achieves scores of zero but BCO+CGL can achieve non-zero scores. With all 300 minutes of human demonstration data, BCO+CGL outperforms BCO on 12 out of 20 games with an average improvement of \textbf{343.6\%}. 
The largest improvement comes from the game Centipede, which increases the average improvement by a large margin (Table \ref{tab:bco_300}). However, %BCO is more challenging than BC. W
we found that BCO is unable to learn an accurate inverse dynamics model for up to seven of the 20 games. For these games, the baseline BCO policy model scores zero and the utilization of gaze is also unable to overcome the shortcomings of the inverse dynamics model. We utilize a publicly available implementation of BCO by \citet{brown2019extrapolating} and do not tune any parameters specifically for the Atari-HEAD dataset \citep{zhang2019atari} for this work.

\subsubsection{CGL Provides Stronger Guidance than GMD}
We test GMD and CGL with BCO and find that on average across 20 games, CGL outperforms GMD both with 15 minutes and 300 minutes of demonstration data (Fig.~\ref{fig:bco_improv}). \citet{chen2019gaze} propose using a uniform dropout probability of 0.7, whereas we test GMD with nine values similar to the hyperparameters tested with CGL -- 0.001, 0.005, 0.01, 0.05, 0.1, 0.3, 0.5, 0.7, 0.9. GMD outperforms CGL in 3 out of 20 games with 15 minutes of demonstration data and 4 out of 20 games with 300 minutes of demonstration data.
One may expect that convolutional dropout helps generalization by reducing over-fitting. However, it is less advantageous since the shared-filter and local-connectivity architecture in convolutional layers is a drastic reduction in the number of parameters and this already reduces the possibility to overfit \cite{hinton2012improving}. Empirical results by Wu et al. \cite{wu2015towards} confirm that the improvement in generalization to test data from convolutional dropout is often inferior to max-pooling or fully-connected dropout.

\subsubsection{CGL Provides Stronger Guidance than Implicit Motion Information}
\begin{figure*}
\centering
\subfigure[An input frame stack for Ms. Pacman]{
\includegraphics[width=0.64\textwidth]{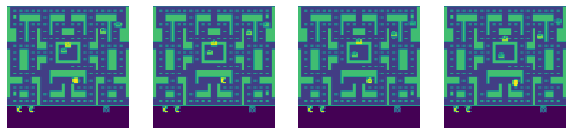}
}
\subfigure[Motion heatmap]{
\includegraphics[width=0.15\textwidth]{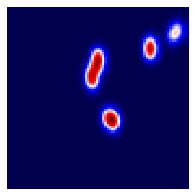}
}
\subfigure[Gaze heatmap]{
\includegraphics[width=0.15\textwidth]{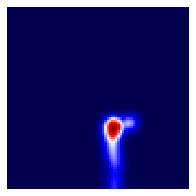}
}
\subfigure[An input frame stack for Asterix]{
\includegraphics[width=0.64\textwidth]{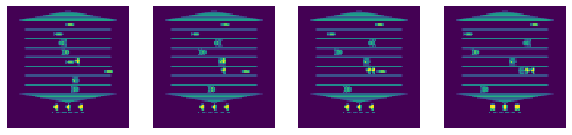}
}
\subfigure[Motion heatmap]{
\includegraphics[width=0.15\textwidth]{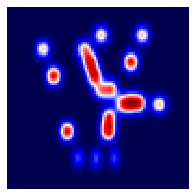}
}
\subfigure[Gaze heatmap]{
\includegraphics[width=0.15\textwidth]{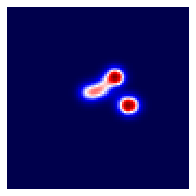}
}
\caption{Motion in the visual game state, i.e. the difference between the last and first frame in an input image stack, cannot alone explain human attention. This is further highlighted with minimal performance gains when CGL utilizes the motion heatmap instead of the human gaze heatmap.}
\label{fig:motion}
\end{figure*}

Prior work has established that human gaze encodes attention which is different from salient regions in a scene (such as motion)~\cite{land2009vision,hayhoe2005eye,rothkopf2007task,tatler2011eye}.
We test whether our proposed approach extracts the additional information from human gaze data, in comparison to what might already be encoded in the visual game state, such as motion. We replace the gaze heat maps used by CGL with heatmaps representing the normalized motion in an input image frame stack (the difference between the last and first frame in a stack shown in Fig.~\ref{fig:motion}). We test this motion-based loss for BCO with the same hyperparameter values as that for CGL. While motion information is also beneficial to improve performance for BCO and is also beneficial in predicting human gaze \citep{zhang2018agil}, the average gain from CGL is higher than that from motion, more so with 300 minutes of demonstration data (Fig.~\ref{fig:bco_improv}). The games Berzerk, Centipede, and River Raid show consistent performance improvements with CGL compared to motion, regardless of the amount of training data used. 
Prior work \cite{yuezhang2018initial} has also shown that using optical flow between two frames as the attention map provides moderate performance improvements in Atari Games. Incorporating both gaze and motion information simultaneously in an auxiliary loss can be investigated as part of future work.

\subsection{CGL Improves T-REX}
T-REX is an inverse reinforcement learning algorithm which compares pairs of trajectory snippet to learn the reward. Along with full 300 minutes of demonstration data, we evaluate with 30 minutes of demonstration data to compare trajectories from two different demonstrations (each demonstration in Atari-HEAD \citep{zhang2019atari} was at least 15 minutes long and T-REX requires at least two demonstrations to compute the reward function). The hyperparameter  $\alpha$ is tuned using a grid search from a set of 9 values --- 0.001, 0.005, 0.01, 0.05, 0.1, 0.3, 0.5, 0.7, 0.9 and the Adam optimizer \cite{kingma2014adam} is used. T-REX+CGL outperforms basic T-REX on 15 out of 20 games both with 30 minutes and 300 minutes worth of training data (Tables \ref{tab:trex_30} and \ref{tab:trex_300}).   
On average, the improvement due to CGL is \textbf{390.4\%} with 30 minutes of data, and \textbf{373.6\%} with 300 minutes of data (Table~\ref{tab:trex_improv}). There are four games that T-REX achieves scores of zero but T-REX+CGL can achieve non-zero scores (Tables \ref{tab:trex_30} and \ref{tab:trex_300}). To the best of our knowledge, CGL is the first method to augment the learning of an IRL algorithm with human gaze.

\subsection{Best Performing Models for each Game}
We then summarize the best game scores obtained from various algorithms presented above. The results are shown in Table~\ref{tab:best}. We notice that CGL augmented methods achieve the best results in 15 out of 20 games. For comparison, we also show DQN scores~\cite{mnih2015human,hessel2018rainbow} as a reference (the evaluation methods are slightly different). With human gaze information (especially with CGL), imitation learning algorithms start to match and even outperform DQN. Note that DQN is trained with 200M samples per game, while IL methods are at most trained with 360K samples (300 minutes of human data). 

\begin{table}[]
    \caption{Average (across 20 games) improvement over the T-REX baseline. Result is presented as as mean$\pm$standard error of the mean (N=20). Individual game scores of all agents can be found in Tables \ref{tab:trex_30} and \ref{tab:trex_300}.
    }
    \centering
    \begin{tabular}{c|c}
    \toprule
Improvement over T-REX (\%) &	T-REX+CGL \\
\midrule
30min data	& $390.4\pm203.5$ \\
\midrule
300min data	& $373.6\pm206.5$ \\
\bottomrule
\end{tabular}
\label{tab:trex_improv}
\end{table}

\begin{table}[]
\caption{A summary of the best game scores obtained. DQN scores are from no-op starts evaluation regime table of~\cite{hessel2017rainbow} , except for game Riverraid~\cite{mnih2015human}. With human gaze information (especially with CGL), imitation learning algorithms start to match and even outperform DQN.}
\resizebox{\columnwidth}{!}{
\begin{tabular}{c|ccc}
\toprule
Game & Algorithm (\#demo)         & Score & DQN Score \\
\midrule
alien                          & AGIL (300min)     & \textbf{2104.7}           & 1620.0                          \\
{\color[HTML]{24292E} asterix} & T-REX+CGL (30min) & \textbf{66445.0}            & 4359.0                          \\
bank\_heist                    & BC-2ch (300min)   & 174.3                     & \textbf{455.0}                  \\
berzerk                        & BCO+CGL (15min)   & \textbf{687.67}           & 585.6                         \\
breakout                       & T-REX+CGL (300min) & \textbf{438.4}           & 385.5                         \\
centipede                      & T-REX+CGL (30min) & \textbf{20762.5}         & 4657.7                        \\
demon\_attack                  & T-REX+CGL (300min)     & \textbf{17589.0}                    & 12149.4             \\
enduro                         & BC+CGL (300min)   & 445.1                     & \textbf{729.0}                  \\
freeway                        & BC-2ch (300min)   & \textbf{31.4}             & 30.8                          \\
frostbite                      & BC+CGL (30min)    & \textbf{5897.7}           & 797.4                         \\
hero                           & BC+CGL (15min)    & 19023.2                   & \textbf{20437.8}              \\
montezuma             & BC+CGL (300min)   & \textbf{1720.0}             & 0.0                             \\
ms\_pacman                     & BC+CGL (30min)    & 2739.7                    & \textbf{3085.6}               \\
name\_this\_game               & AGIL (300min)     & 5817.0                      & \textbf{8207.8}               \\
phoenix                        & AGIL (300min)     & 5140.0                      & \textbf{8485.2}               \\
riverraid                      & T-REX+CGL (300min) & 7370.0                   & \textbf{8316.0}                 \\
road\_runner                   & BC+CGL (300min)   & 33510.0                     & \textbf{39544.0}                \\
seaquest                       & T-REX+CGL (30min) & 759.3                   & \textbf{5860.6}               \\
space\_invaders                & T-REX+CGL (300min) & 1563.7                     & \textbf{1692.3}               \\
venture                        & BC+CGL (15min)    & \textbf{376.7}            & 163.0 \\
\bottomrule
\end{tabular}}
\label{tab:best}
\end{table}

\subsection{Visualizing CGL Attention} We can analyze whether the CGL agents have successfully learned to pay attention to the critical regions highlighted by human saliency maps in two ways. First, we directly visualize the activation map of the networks trained with and without CGL, which has already been shown in Fig.~\ref{fig:activations} for a trained BCO agent. However, this only shows that the convolutional layer we applied CGL to behaves as expected.

\begin{figure}
\centering
\subfigure[Game state]{
\includegraphics[width=0.11\textwidth,height=0.11\textwidth]{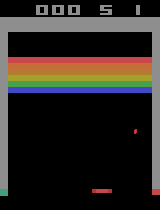}
}
\hspace{-2.8mm}
\subfigure[Human]{
\includegraphics[width=0.11\textwidth]{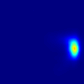}
}
\hspace{-2.8mm}
\subfigure[T-REX]{
\includegraphics[width=0.11\textwidth]{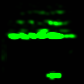}
 }
 \hspace{-2.7mm}
 \subfigure[T-REX+CGL]{
\includegraphics[width=0.11\textwidth]{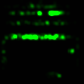}
 }

\subfigure[Game state]{
\includegraphics[width=0.11\textwidth,height=0.11\textwidth]{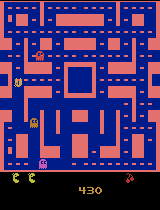}
}
\hspace{-2.8mm}
\subfigure[Human]{
\includegraphics[width=0.11\textwidth]{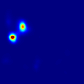}
}
\hspace{-2.8mm}
\subfigure[T-REX]{
\includegraphics[width=0.11\textwidth]{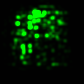}
 }
 \hspace{-2.7mm}
 \subfigure[T-REX+CGL]{
\includegraphics[width=0.11\textwidth]{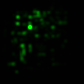}
 }
\caption{CGL guides the T-REX reward network (d) to focus on parts of the state space which the human attends to (b) for Breakout (a). A failure case happens where CGL is unable to guide the T-REX reward network (h) to attend to both modes which the human attends to (f) for MsPacman (g). The attention maps of deep networks are generated using the method proposed by~\citet{greydanus2018visualizing}. 
}
\label{fig:greydanus}
\vspace{-3mm}
\end{figure}

We use a second method to show that the whole trained network has learned to attend to the desired region with CGL. We visualize the attention maps of trained CGL agents with a method commonly used to provide visual interpretations of deep RL agents~\cite{greydanus2018visualizing}. The algorithm takes an input image $I$ and applies a Gaussian filter to a pixel location $(i,j)$ to blur the image. This manipulation adds spatial uncertainty to the surrounding region and produces a perturbed image $\Phi(I,i,j)$. A saliency score for this pixel $(i,j)$ can be defined as how much the blurred image changes the network output~\cite{greydanus2018visualizing}. Doing this for every pixel results in a saliency map that approximates the ``attention" of a network. The results for a case of Breakout where CGL outperforms the baseline T-REX method can be found in Fig.~\ref{fig:greydanus}, where the T-REX+CGL agent successfully learned to focus on the ball like the human did, while the T-REX agent did not. We also highlight a failure case with MsPacman, where the CGL agent does not outperform baseline T-REX. We find that the network fails to attend to both modes of attention in the human's gaze map.

\subsection{Reducing Causal Confusion with CGL}
\label{sec:causal}
Discriminative models for IL such as BC are non-causal, i.e. the training procedure is unaware of the causal structure of the interaction between the demonstrator and the environment.
Causal misidentification is the phenomenon where cloned policies fail by misidentifying the causes of the demonstrator's actions.
A very problematic effect of distributional shift in BC can lead to causal misidentification.
This is exacerbated by the causal structure of sequential action: the very fact that current actions cause future observations often introduces complex new nuisance correlates.

\begin{figure}
\centering
\subfigure[Breakout]{
\includegraphics[width=0.11\textwidth,height=0.11\textwidth]{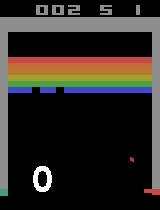}
}
\hspace{-2.8mm}
\subfigure[Asterix]{
\includegraphics[width=0.11\textwidth,height=0.11\textwidth]{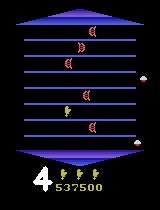}
}
\hspace{-2.8mm}
\subfigure[Demon Attack]{
\includegraphics[width=0.11\textwidth,height=0.11\textwidth]{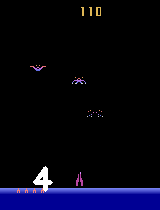}
 }
 \hspace{-2.7mm}
 \subfigure[Freeway]{
\includegraphics[width=0.11\textwidth,height=0.11\textwidth]{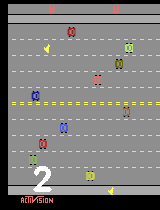}
 }
\caption{Confounded states with past actions (text indicating past action identifiers is superimposed on the visual state) to test reduction of causal confusion with CGL. The design follows methodology proposed by~\citet{de2019causal}.}
\label{fig:confounded}
\vspace{-3mm}
\end{figure}

Prior work on understanding causal confusion in IL \citep{de2019causal} uses past action information (often correlated with current action) to identify if the IL algorithm is in a causal confusion trap.  
To understand the performance gains of CGL, we investigate if it disambiguates the intent of the user in the demonstrated actions by eliminating causal confusion.
We overlay the four frame image stack (state) with actions from the last frame in the previous stack (Fig. \ref{fig:confounded}). This lays a causal confusion trap for the IL agent. If the agent can ignore the new correlated action information that is part of the state space, it hints towards the fact that the agent learns to ignore those features and perform better empirically. We find that on average (across 20 games), CGL agents suffer less when trained with confounded data compared to the BC baseline (\textbf{-34.0\%} versus \textbf{-47.8\%}). Moreover, when trained with confounded data, BC+CGL outperforms BC trained with confounded data by \textbf{571\%} %and still perform better than BC agents 
(Table \ref{tab:bc_confounded}), indicating that guidance with human gaze via CGL greatly reduces the causal confusion introduced with overlaid past action information. 
This hints to the fact that in addition to directing the attention of the network to learn a better mapping between states and actions, part of the gains from using human gaze data with CGL can come from reducing causal confusion.

\section{Conclusions}

In this work, we introduced an auxiliary coverage-based gaze loss (CGL) term which guides the training of any imitation learning network with convolutional layers. Our experiments showed improved performance on several Atari games over standard imitation learning algorithms. Our approach provides these gains without requiring gaze prediction at test time or increasing the model complexity of existing algorithms. We outperform a baseline method (GMD), which also does not increase model complexity. CGL is more efficient in terms of both learnable parameters and data efficiency when compared to a state-of-the-art gaze-augmented IL method (AGIL) which utilizes gaze in the form of additional input to a BC algorithm. AGIL requires gaze prediction at test time and is shown to gain performance by increasing model complexity alone. Our approach improved performance by utilizing gaze without these shortcomings. We also highlighted that utilizing human gaze provides additional information to what is encoded implicitly in the game state (such as motion). Our work confirms prior research showing gaze can help extract more information from a demonstrator than traditional state-action pairs, bridging some of the gap in performance between IL and RL agents. 

\section{Discussion and Future Work}

Human attention can be seen as a form of spatial prior on the visual input. In deep learning research, this prior is often used as a mask to filter out unimportant information (e.g., AGIL). This approach has two main drawbacks. First, it requires the mask at testing time. Secondly, some unattended visual features handled by human memory systems could still be useful for decisions. Therefore, completely filtering out all unattended information seems inappropriate. In this work, we present a novel method to incorporate gaze effectively that only requires access to human gaze data at training time. Moreover, by utilizing a coverage-based loss, this method highlights the attended features while keeping the unattended features available for the learning agent. This novel method in training deep neural networks can be applied to other learning tasks that utilize other forms of spatial priors. 

One limitation of our approach is the need for extensive hyperparameter tuning to balance the linear combination of loss functions during training. Moreover, finding good solutions to optimally balance multi-objective losses can be especially challenging if the Pareto front of the loss landscape is concave \citep{boyd2004convex}. Often the shape of the Pareto front for neural network architectures is unknown and it can be hard to find an optimal solution to minimize multi-objective loss functions. \citet{platt1988constrained} proposed a theoretical framework (a modified differential method of multipliers) to tune the balance between the losses in a semantically useful way using stochastic gradient descent, no matter the shape of the invisible Pareto front. They introduce an additional damping hyper-parameter which trades the time to find the Pareto front with the time to converge to a solution on that front. 
Going forward, we envision that utilizing such theoretical findings \citep{bworld} can help the training of CGL based deep networks
by potentially reducing the effort for tuning hyper-parameters and further improving the performance gains of incorporating gaze. 

Additionally, human gaze and actions from demonstrations may be correlated in time. Our approach only utilizes gaze per game state, and so do all other approaches we compare against. Utilizing temporal connections in the gaze signal is a direction for future work. 
We hope our work encourages the research community to innovate on other novel ideas for efficiently incorporating human attention as part of different learning frameworks.

% \input{6-appendix}

%%%%%%%%%%%%%%%%%%%%%%%%%%%%%%%%%%%%%%%%%%%%%%%%%%%%%%%%%%%%%%%%%%%%%%%%

%%% The acknowledgments section is defined using the "acks" environment
%%% (rather than an unnumbered section). The use of this environment 
%%% ensures the proper identification of the section in the article 
%%% metadata as well as the consistent spelling of the heading.

% \begin{comment}
\begin{acks}
This work has taken place in the Personal Autonomous Robotics Lab (PeARL) at The University of Texas at Austin. PeARL research is supported in part by the NSF (IIS-1724157, IIS-1638107, IIS-1749204, IIS-1925082) and ONR (N00014-18-2243).  This research was also sponsored by the Army Research Office and was accomplished under Cooperative Agreement Number W911NF-19-2-0333. The views and conclusions contained in this document are those of the authors and should not be interpreted as representing the official policies, either expressed or implied, of the Army Research Office or the U.S. Government. The U.S. Government is authorized to reproduce and distribute reprints for Government purposes notwithstanding any copyright notation herein. 
This work is also partially supported by NIH (EY05729).
\end{acks}
% \end{comment}

%%%%%%%%%%%%%%%%%%%%%%%%%%%%%%%%%%%%%%%%%%%%%%%%%%%%%%%%%%%%%%%%%%%%%%%%

%%% The next two lines define, first, the bibliography style to be 
%%% applied, and, second, the bibliography file to be used.

\bibliographystyle{ACM-Reference-Format} 
\bibliography{sample}

%%%%%%%%%%%%%%%%%%%%%%%%%%%%%%%%%%%%%%%%%%%%%%%%%%%%%%%%%%%%%%%%%%%%%%%%
\clearpage

\appendix
\section{Demonstration Data}

We use demonstrations collected in the Arcade Learning Environment (ALE) \cite{bellemare2013arcade} for a total of 20 games. Demonstrations from four participants are publicly available as part of the Atari-HEAD dataset \cite{zhang2019atari}. The participants were only allowed to play for 15 minutes and were required to rest for at least 5 minutes before the next trial. The trial IDs corresponding to 15 minutes and 30 minutes of human demonstration data are listed in Tables \ref{tab:15} and \ref{tab:30} respectively.

Gaze data was recorded using an EyeLink $1000$ eye tracker at $1000$ Hz. The game screen was $64.6 \times 40.0$ cm (or $1280\times840$ in pixels), and the distance to the subjects' eyes was $78.7$ cm. The visual angle of the screen was $44.6 \times 28.5$ visual degrees, where the visual angle of an object is a measure of the size of the object's image on the retina. In the default ALE setting, the game runs continuously at $60$ Hz, a speed that is very challenging even for expert human players. An innovative feature of the Atari-HEAD \cite{zhang2019atari} setup is that the game pauses at every frame, until a keyboard action is taken by the human player. This allows users to fixate at all critical locations of the state space before taking an action, giving enough reaction time to a participant and producing a richer gaze signal at every time step. If desired, the subjects can hold down a key and the game will run continuously at 20Hz, a speed that is reported to be comfortable for most players.

\subsection{Human Gaze Models}
 Since our baseline methods require gaze estimates at test time, we use gaze heatmaps from a prediction network for consistency across all methods. Gaze data collected with demonstrations is used to generate gaze heatmaps, which are in turn used for training gaze prediction networks with the four frames of image stack representing the game state, optical flow and saliency maps as input. We use convolution-deconvolution networks trained on real human gaze data, following the architecture by \citet{zhang2018agil} for gaze prediction. For experiments with 15 minutes of demonstration data, we use the gaze data corresponding to the same set of demonstrations to train the gaze prediction network. We follow a similar approach for experiments with 30 minutes and 300 minutes of demonstration data. To generate gaze heatmaps from the data collected during demonstrations, the discrete gaze positions are converted into a continuous distribution \cite{bylinskii2018different} by blurring each fixation location using a Gaussian with a standard deviation equal to one visual degree \cite{le2013methods}.

\begin{table}[]
\caption{The human demonstration data used for BC and BCO (15 minutes) experiments. The trial IDs are human data trial IDs from Atari-HEAD dataset, each trial contains 15 minutes of human demonstration data. For experiments with 300 minutes of data, we include all human trials.}
\begin{tabular}{c|c}
\toprule
Game & Trial ID \\
\midrule
alien                          & 474   \\
asterix &                     213   \\
bank\_heist                    & 456   \\
berzerk                        & 468   \\
breakout                       & 198   \\
centipede                      & 204   \\
demon\_attack                  & 475   \\
enduro                         & 473   \\
freeway                        & 616   \\
frostbite                      & 479   \\
hero                           & 443   \\
montezuma\_revenge             & 480   \\
ms\_pacman                     & 199   \\
name\_this\_game               & 484   \\
phoenix                        & 180   \\
riverraid                      & 463   \\
road\_runner                   & 483   \\
seaquest                       & 185   \\
space\_invaders                & 455   \\
venture                        & 486  \\
\bottomrule
\end{tabular}
\label{tab:15}
\end{table}

\begin{table}[]
\caption{The human demonstration data used for TREX (30 minutes) experiments. The trial IDs are human data trial IDs from Atari-HEAD dataset, each trial contains 15 minutes of human demonstration data. For experiments with 300 minutes of data, we include all human trials.}
\begin{tabular}{c|c}
\toprule
Game & Trial IDs \\
\midrule
alien                          & 464, 474   \\
asterix &                     213, 301   \\
bank\_heist                    & 438, 456   \\
berzerk                        & 454, 468   \\
breakout                       & 198, 218   \\
centipede                      & 204, 210   \\
demon\_attack                  & 445, 475   \\
enduro                         & 453, 473   \\
freeway                        & 616, 617   \\
frostbite                      & 466, 479   \\
hero                           & 431, 443   \\
montezuma\_revenge             & 469, 480   \\
ms\_pacman                     & 199, 209   \\
name\_this\_game               & 458, 484   \\
phoenix                        & 180, 408   \\
riverraid                      & 444, 463   \\
road\_runner                   & 460, 483   \\
seaquest                       & 185, 212   \\
space\_invaders                & 364, 455   \\
venture                        & 471, 486  \\
\bottomrule
\end{tabular}
\label{tab:30}
\end{table}

\section{Individual Game Scores}

The individual game scores corresponding to the aggregated results shown in Fig. \ref{fig:bc_improv}, Fig. \ref{fig:bco_improv} and Table \ref{tab:trex_improv} are listed below.

\begin{table*}[hbt!]
\caption{Game scores obtained when using 15 minutes of human demonstration data to train the agents. Results are presented as mean$\pm$standard error of the mean (N=30). The agents we compare are behavioral cloning agent (BC), two channeled behavioral cloning agent (BC-2ch), Attention-guided imitation learning agent (AGIL), and proposed coverage-based loss agent (CGL). The improvement columns show the relative improvement over the BC baseline. Average results over all 20 games are presented in Fig. \ref{fig:bc_improv}.}
\resizebox{\textwidth}{!}{
\begin{tabular}{c|ccccccc}
\toprule
& BC & BC-2ch         & AGIL                   & BC+CGL  & Improv-BC-2ch & Improv-AGIL & Improv-CGL \\
\midrule
alien                          & 1575$\pm$176.8             & 1296.7$\pm$140.8  & 1866.7$\pm$171.3          & \textbf{2044.7$\pm$242.1}  & -17.7\%                           & 18.5\%                          & 29.8\%                         \\
{\color[HTML]{24292E} asterix} & 285$\pm$28.2               & 283.3$\pm$31.1    & 275$\pm$35.9              & \textbf{426.7$\pm$27.8}    & -0.6\%                            & -3.5\%                          & 49.7\%                         \\
bank\_heist                    & 86.3$\pm$9.0               & 129.3$\pm$17.7    & \textbf{169$\pm$13.1}     & 143$\pm$14.8               & 49.8\%                            & 95.8\%                          & 65.7\%                         \\
berzerk                        & 330.7$\pm$22.0             & 350$\pm$22.5      & 251.7$\pm$19.1            & \textbf{366.7$\pm$19}      & 5.8\%                             & -23.9\%                         & 10.9\%                         \\
breakout                       & 2.2$\pm$0.3                & 2.7$\pm$0.3       & \textbf{4.9$\pm$0.4}      & 3.7$\pm$0.4                & 22.7\%                            & 122.7\%                         & 68.2\%                         \\
centipede                      & 4378.8$\pm$442.6           & 5762.1$\pm$687.6  & 4600.2$\pm$333.8          & \textbf{6075.9$\pm$845.1}  & 31.6\%                            & 5.1\%                           & 38.8\%                         \\
demon\_attack                  & 112.2$\pm$13.9             & 177$\pm$18.7      & 148$\pm$16.9              & \textbf{205.2$\pm$41.9}    & 57.8\%                            & 31.9\%                          & 82.9\%                         \\
enduro                         & \textbf{11.7$\pm$2.0}      & 7.8$\pm$1.4       & 0$\pm$0                   & 4.2$\pm$1.4                & -33.3\%                           & -100.0\%                        & -64.1\%                        \\
freeway                        & 29.4$\pm$0.2               & 28.5$\pm$0.3      & 28.1$\pm$0.3              & \textbf{30$\pm$0.3}        & -3.1\%                            & -4.4\%                          & 2.0\%                          \\
frostbite                      & 1628.3$\pm$246.4           & 1406.3$\pm$266.5  & \textbf{3185$\pm$352.9}   & 2973$\pm$279               & -13.6\%                           & 95.6\%                          & 82.6\%                         \\
hero                           & 13255.3$\pm$845.1          & 18877.2$\pm$509.0 & 15582.7$\pm$789.7         & \textbf{19023.2$\pm$679.7} & 42.4\%                            & 17.6\%                          & 43.5\%                         \\
montezuma\_revenge             & 100$\pm$31.6               & 0$\pm$0.0         & 0$\pm$0                   & \textbf{1200$\pm$159.2}    & -100.0\%                          & -100.0\%                        & 1100.0\%                       \\
ms\_pacman                     & 843.3$\pm$62.8             & 783.3$\pm$55.8    & 1072.3$\pm$74.8           & \textbf{1348.3$\pm$206.9}  & -7.1\%                            & 27.2\%                          & 59.9\%                         \\
name\_this\_game               & 1917.3$\pm$130.2           & 2153.3$\pm$169.1  & \textbf{2832$\pm$194.2}   & 2646.3$\pm$156.3           & 12.3\%                            & 47.7\%                          & 38.0\%                         \\
phoenix                        & 1060$\pm$172.4             & 1105$\pm$159.7    & 1171.7$\pm$147.6          & \textbf{2193.7$\pm$200.5}  & 4.2\%                             & 10.5\%                          & 107.0\%                        \\
riverraid                      & 2771.7$\pm$141.8           & 2701.3$\pm$128.0  & \textbf{3900.3$\pm$223.6} & 2965.3$\pm$184.8           & -2.5\%                            & 40.7\%                          & 7.0\%                          \\
road\_runner                   & 7840$\pm$553.3             & 3820$\pm$372.3    & 6920$\pm$518.8            & \textbf{12723.3$\pm$376.7} & -51.3\%                           & -11.7\%                         & 62.3\%                         \\
seaquest                       & 194$\pm$11.4               & 162$\pm$9.8       & 198$\pm$12.7              & \textbf{216$\pm$11.2}      & -16.5\%                           & 2.1\%                           & 11.3\%                         \\
space\_invaders                & 275$\pm$26.2               & 275$\pm$29.3      & 254.7$\pm$21.1            & \textbf{314$\pm$26}        & 0.0\%                             & -7.4\%                          & 14.2\%                         \\
venture                        & 196.7$\pm$26.5             & 273.3$\pm$27.1    & 73.3$\pm$24.9             & \textbf{376.7$\pm$16.1}    & 38.9\%                            & -62.7\%                         & 91.5\%                         \\
\bottomrule
average                        & -  & -              & -                      & -                       & 1.0\%    
& 10.1\%                          & \textbf{95.1\%} \\
\bottomrule
\end{tabular}}
\label{tab:bc_15}
\end{table*}

\begin{table*}[]
\caption{Game scores obtained when using all 300 minutes of human demonstration data to train the agents. Results are presented as mean$\pm$standard error of the mean (N=30). The agents we compare are behavioral cloning agent (BC), two channeled behavioral cloning agent (BC-2ch), Attention-guided imitation learning agent (AGIL), and proposed coverage-based loss agent (CGL). The improvement columns show the relative improvement over the BC baseline. Average results are shown in Fig. \ref{fig:bc_improv}.}
\resizebox{\textwidth}{!}{
\begin{tabular}{cccccccc}
\toprule
& BC & BC-2ch         & AGIL                     & BC+CGL                 & Improv-BC-2ch & Improv-AGIL & Improv-CGL \\
\midrule
alien                          & 694$\pm$97.5       & 1303.7$\pm$147.6        & \textbf{2104.7$\pm$180.2} & 2027.3$\pm$140.1            & 87.9\%                            & 203.3\%                         & 192.1\%                        \\
asterix & 516.7$\pm$54.9     & 465$\pm$32              & 455$\pm$50.1              & \textbf{773.3$\pm$63.4}     & -10.0\%                           & -11.9\%                         & 49.7\%                         \\
bank\_heist                    & 102.3$\pm$8        & \textbf{174.3$\pm$13.9} & 117.3$\pm$13.3            & 156.3$\pm$12.5              & 70.4\%                            & 14.7\%                          & 52.8\%                         \\
berzerk                        & 256.7$\pm$17.4     & 379.3$\pm$29.3          & 188.3$\pm$24.5            & \textbf{435$\pm$42.3}       & 47.8\%                            & -26.6\%                         & 69.5\%                         \\
breakout                       & 1.7$\pm$0.3        & \textbf{3.6$\pm$0.4}    & \textbf{3.6$\pm$0.4}      & 2.9$\pm$0.3                 & 111.8\%                           & 111.8\%                         & 70.6\%                         \\
centipede                      & 7704.5$\pm$967     & 7856.8$\pm$903.1        & 9073.7$\pm$914.8          & \textbf{9330$\pm$922}       & 2.0\%                             & 17.8\%                          & 21.1\%                         \\
demon\_attack                  & 848.2$\pm$140.3    & 333.8$\pm$38.6          & \textbf{2156.2$\pm$295.2} & 1375$\pm$216.7              & -60.6\%                           & 154.2\%                         & 62.1\%                         \\
enduro                         & 386$\pm$12.1       & 385.4$\pm$12            & 278.7$\pm$17.5            & \textbf{445.1$\pm$17}       & -0.2\%                            & -27.8\%                         & 15.3\%                         \\
freeway                        & 27.6$\pm$0.3       & \textbf{31.4$\pm$0.1}   & 28.9$\pm$0.3              & 30.2$\pm$0.2                & 13.8\%                            & 4.7\%                           & 9.4\%                          \\
frostbite                      & 2016.7$\pm$180.2   & 2331.7$\pm$240.4        & 1980$\pm$176.4            & \textbf{3253$\pm$254.5}     & 15.6\%                            & -1.8\%                          & 61.3\%                         \\
hero                           & 9519.7$\pm$874.9   & 11152$\pm$1079.9        & 7685.7$\pm$1100.6         & \textbf{16936.5$\pm$1342.6} & 17.1\%                            & -19.3\%                         & 77.9\%                         \\
montezuma\_revenge             & 490$\pm$109        & 1480$\pm$168.8          & 553.3$\pm$70.3            & \textbf{1720$\pm$156.3}     & 202.0\%                           & 12.9\%                          & 251.0\%                        \\
ms\_pacman                     & 1200$\pm$123.2     & 1511$\pm$144.1          & 1272.3$\pm$128.6          & \textbf{1590$\pm$197.1}     & 25.9\%                            & 6.0\%                           & 32.5\%                         \\
name\_this\_game               & 2887$\pm$177.8     & 5732$\pm$288.2          & \textbf{5817$\pm$341.7}   & 5405.3$\pm$267              & 98.5\%                            & 101.5\%                         & 87.2\%                         \\
phoenix                        & 4029$\pm$279.8     & 4597.3$\pm$324.1        & \textbf{5140$\pm$405.2}   & 4472.7$\pm$440.2            & 14.1\%                            & 27.6\%                          & 11.0\%                         \\
riverraid                      & 2806$\pm$164.5     & 2266.3$\pm$65.7         & 2555.7$\pm$56.2           & \textbf{3452.7$\pm$242.1}   & -19.2\%                           & -8.9\%                          & 23.0\%                         \\
road\_runner                   & 29433.3$\pm$1230.3 & 31776.7$\pm$1426.2      & 29410$\pm$1516.2          & \textbf{33510$\pm$1026.3}   & 8.0\%                             & -0.1\%                          & 13.9\%                         \\
seaquest                       & 175.5$\pm$30.7     & 182.1$\pm$12.4          & 443$\pm$124.6             & \textbf{610.7$\pm$106.8}    & 3.8\%                             & 152.4\%                         & 248.0\%                        \\
space\_invaders                & 243.8$\pm$26.7     & 196$\pm$26.4            & 206$\pm$23                & \textbf{363.5$\pm$35.7}     & -19.6\%                           & -15.5\%                         & 49.1\%                         \\
venture                        & 73.3$\pm$27.9      & 43.3$\pm$20.9           & \textbf{330$\pm$21.7}     & 313.3$\pm$29                & -40.9\%                           & 350.2\%                         & 327.4\%                        \\
\bottomrule
average &         -        &                -      &            -            &             -             & 28.4\%                            & 52.3\%                          & \textbf{86.2\%}  \\             
\bottomrule
\end{tabular}}
\label{tab:bc_300}
\end{table*}

\begin{table*}[]
\caption{Game scores obtained when using 15 minutes of human demonstration data to train the agents. Results are presented as mean$\pm$standard error of the mean (N=30). The agents we compare are behavioral cloning from observation agent (BCO), gaze-modulated dropout (GMD), BCO with motion information, and BCO+CGL. The improvement columns show the relative improvement over the BCO baseline. ``-" indicates that the baseline score is zero hence the relative improvement is not calculated and is not counted in the average. Average results over all 20 games are presented in Fig. \ref{fig:bco_improv}.}
\resizebox{\textwidth}{!}{
\begin{tabular}{c|ccccccc}
\toprule
                               & BCO              & BCO+GMD                       & BCO+Motion                    & BCO+CGL                     & Improv-GMD & Improv-Motion & Improvement-CGL \\
\midrule
alien                          & 0.0 $\pm$ 0.0      & \textbf{140.00 $\pm$ 0.00}  & \textbf{140.00 $\pm$ 0.00}  & \textbf{140.00 $\pm$ 0.00}    & -         & -             & -               \\
asterix & 288.33 $\pm$ 30.48 & 645.00 $\pm$ 33.28          & \textbf{700.00 $\pm$ 0.00}  & 253.33 $\pm$ 9.11             & 123.70\%                       & 142.80\%                          & -12.10\%                            \\
bank\_heist                    & 0.0 $\pm$ 0.0      & 0.00 $\pm$ 0.00             & \textbf{8.67 $\pm$ 0.62}    & 0.00 $\pm$ 0.00               & 0\%                               & -             & 0\%                                    \\
berzerk                        & 158.33 $\pm$ 16.45 & 263.67 $\pm$ 26.36          & 528.33 $\pm$ 17.11          & \textbf{687.67 $\pm$ 27.98}   & 66.50\%                        & 233.70\%                          & 334.30\%                            \\
breakout                       & 0.0 $\pm$ 0.0      & \textbf{2.30 $\pm$ 0.08}    & \textbf{2.30 $\pm$ 0.08}    & 0.60 $\pm$ 0.17               & -          & -             & -               \\
centipede                      & 646.83 $\pm$ 79.70 & 2707.83 $\pm$ 309.30        & 3234.13 $\pm$ 167.51        & \textbf{5047.93 $\pm$ 410.74} & 318.60\%                       & 400\%                             & 680.41\%                            \\
demon\_attack                  & 157.33 $\pm$ 20.87 & \textbf{844.00 $\pm$ 88.40} & 806.00 $\pm$ 85.25          & 791.83 $\pm$ 77.56            & 436.50\%                       & 412.30\%                          & 403.30\%                            \\
enduro                         & 0.0 $\pm$ 0.0      & 0.13 $\pm$ 0.13             & 1.57 $\pm$ 0.72             & \textbf{7.77 $\pm$ 0.76}      & -          & -             & -               \\
freeway                        & 0.0 $\pm$ 0.0      & \textbf{21.30 $\pm$ 0.21}   & \textbf{21.30 $\pm$ 0.21}   & \textbf{21.30 $\pm$ 0.21}     & -          & -             & -               \\
frostbite                      & 116.33 $\pm$ 6.86  & 79.33 $\pm$ 2.94            & 80.67 $\pm$ 6.43            & \textbf{160.00 $\pm$ 0.00}    & -31.80\%                       & -30.70\%                          & 37.50\%                             \\
hero                           & 0.0 $\pm$ 0.0      & 0.00 $\pm$ 0.00             & 0.00 $\pm$ 0.00             & 0.00 $\pm$ 0.00               & 0\%                               & 0\%                                & 0\%                                  \\
montezuma\_revenge             & 0.0 $\pm$ 0.0      & 0.00 $\pm$ 0.00             & 0.00 $\pm$ 0.00             & 0.00 $\pm$ 0.00               & 0\%                               & 0\%                                & 0\%                                 \\
ms\_pacman                     & 60 $\pm$ 0         & \textbf{379.00 $\pm$ 11.27} & 210.00 $\pm$ 0.00           & 210.00 $\pm$ 0.00             & 531.70\%                       & 250\%                             & 250\%                               \\
name\_this\_game               & 694.67 $\pm$ 86.88 & 2183.00 $\pm$ 52.93         & \textbf{2770.00 $\pm$ 0.00} & \textbf{2770.00 $\pm$ 0.00}   & 214.20\%                       & 298.80\%                          & 298.80\%                            \\
phoenix                        & 282 $\pm$ 28.95    & 352.67 $\pm$ 29.26          & \textbf{407.33 $\pm$ 66.01} & 256.67 $\pm$ 22.43            & 25.10\%                        & 44.40\%                           & -9\%                                \\
riverraid                      & 360 $\pm$ 0        & 1029.33 $\pm$ 13.48         & 236.00 $\pm$ 10.93          & \textbf{1250.00 $\pm$ 0.00}   & 185.90\%                       & -34.40\%                          & 247.20\%                            \\
road\_runner                   & 0.0 $\pm$ 0.0      & 473.33 $\pm$ 77.16          & 500.00 $\pm$ 91.29          & \textbf{956.67 $\pm$ 9.05}    & -          & -             & -               \\
seaquest                       & 0.0 $\pm$ 0.0      & 102.00 $\pm$ 4.46           & \textbf{140.00 $\pm$ 0.00}  & \textbf{140.00 $\pm$ 0.00}    & -          & -             & -               \\
space\_invaders                & 220.83 $\pm$ 27.76 & 195.17 $\pm$ 17.64          & \textbf{285.00 $\pm$ 0.00}  & 270.00 $\pm$ 0.00             & -11.60\%                       & 29.10\%                           & 22.30\%                             \\
venture                        & 0.0 $\pm$ 0.0      & 0.00 $\pm$ 0.00             & 0.00 $\pm$ 0.00             & 0.00 $\pm$ 0.00               & 0\%                               & 0\%                                  & 0\%                                    \\
\bottomrule
average                        & -                &                           &                           & -                           & 132.8\%                        & 134.3\%                           & \textbf{160.9\%}      \\                
\bottomrule
\end{tabular}}
\label{tab:bco_15}
\end{table*}

\begin{table*}[]
\caption{Game scores obtained when using 300 minutes of human demonstration data to train the agents. Results are presented as mean$\pm$standard error of the mean (N=30). The agents we compare are behavioral cloning from observation agent (BCO), gaze-modulated dropout (GMD), BCO with motion information, and BCO+CGL. The improvement columns show the relative improvement over the BCO baseline. ``-" indicates that the baseline score is zero hence the relative improvement is not calculated and is not counted in the average. Average results over all 20 games are presented in Fig. \ref{fig:bco_improv}.}
\resizebox{\textwidth}{!}{
\begin{tabular}{c|ccccccc}
\toprule
                               & BCO              & BCO+GMD                       & BCO+Motion                    & BCO+CGL                     & Improv-GMD & Improv-Motion & Improvement-CGL \\
\midrule
alien                          & 140.00 $\pm$ 0.00           & 140.00 $\pm$ 0.00    & 140.00 $\pm$ 0.00           & 140.00 $\pm$ 0.00             & 0\%                            & 0\%                               & 0\%                            \\
asterix & 181.00 $\pm$ 13.02             & 276.67 $\pm$ 24.24   & 650.00 $\pm$ 0.00           & \textbf{690.00 $\pm$ 27.33}   & 52.90\%                        & 259.10\%                          & 281.20\%                       \\
bank\_heist                    & 0.00 $\pm$ 0.00             & 0.00 $\pm$ 0.00      & 0.00 $\pm$ 0.00             & 0.00 $\pm$ 0.00               & 0\%                            & 0\%                               & 0\%                            \\
berzerk                        & 145.00 $\pm$ 15.87          & 229.00 $\pm$ 22.40   & 539.00 $\pm$ 22.99          & \textbf{572.33 $\pm$ 17.71}   & 57.90\%                        & 271.70\%                          & 294.70\%                       \\
breakout                       & 0.30 $\pm$ 0.08              & \textbf{2.00 $\pm$ 0.00}      & 0.60 $\pm$ 0.17             & 0.17 $\pm$ 0.07               & 566.60\%                       & 100\%                             & -43.30\%                       \\
centipede                      & 184.00 $\pm$ 0.91              & 3309.03 $\pm$ 186.34 & 3668.27 $\pm$ 192.32        & \textbf{8391.03 $\pm$ 557.52} & 1698.40\%                      & 1893.60\%                         & 4460.30\%                      \\
demon\_attack                  & 127.67 $\pm$ 19.33          & 180.00 $\pm$ 18.60   & \textbf{806.00 $\pm$ 85.25}          & \textbf{806.00 $\pm$ 85.25}            & 41.00\%                        & 531.30\%                          & 531.30\%                       \\
enduro                         & \textbf{2.93 $\pm$ 0.81}             & 0.37 $\pm$ 0.17      & 0.07 $\pm$ 0.05             & 0.00 $\pm$ 0.00               & -87.20\%                       & -97.60\%                          & -100\%                         \\
freeway                        & 0.00 $\pm$ 0.00             & \textbf{21.30 $\pm$ 0.21}     & \textbf{21.30 $\pm$ 0.21}            & \textbf{21.30 $\pm$ 0.21}              & -          & -             & -          \\
frostbite                      & 102.33 $\pm$ 6.14           & \textbf{329.67 $\pm$ 53.65}   & 160.00 $\pm$ 0.00  & 124.00 $\pm$ 12.68            & 222.20\%                       & 56.40\%                           & 21.20\%                        \\
hero                           & 0.00 $\pm$ 0.00             & \textbf{150.00 $\pm$ 0.00}    & 0.00 $\pm$ 0.00             & 0.00 $\pm$ 0.00               & -          & 0\%                               & 0\%                            \\
montezuma\_revenge             & 0.00 $\pm$ 0.00             & 0.00 $\pm$ 0.00      & 0.00 $\pm$ 0.00             & 0.00 $\pm$ 0.00               & 0\%                            & 0\%                               & 0\%                            \\
ms\_pacman                     & 60.00 $\pm$ 0                  & \textbf{210.00 $\pm$ 0.00}    & \textbf{210.00 $\pm$ 0.00}           & \textbf{210.00 $\pm$ 0.00}                   & 250.00\%                          & 250.00\%                             & 250.00\%                          \\
name\_this\_game               & 1158.00 $\pm$ 50.88         & 1808.00 $\pm$ 75.44  & \textbf{2770.00 $\pm$ 0.00}          & \textbf{2770.00 $\pm$ 0.00}            & 56.10\%                        & 139.20\%                          & 139.20\%                       \\
phoenix                        & 147.33 $\pm$ 6.22           & 444.00 $\pm$ 52.35   & \textbf{474.00 $\pm$ 75.41} & 356.43 $\pm$ 70.41            & 201.40\%                       & 221.70\%                          & 141.90\%                       \\
riverraid                      & 360.00 $\pm$ 0.00           & \textbf{1646.33 $\pm$ 50.27}  & 440.00 $\pm$ 0.00           & 1222.00 $\pm$ 3.98  & 357.30\%                       & 22.20\%                           & 239.40\%                       \\
road\_runner                   & 0.00 $\pm$ 0.00             & 493.33 $\pm$ 97.52   & \textbf{956.67 $\pm$ 9.05}  & 0.00 $\pm$ 0.00               & -          & -             & 0\%                            \\
seaquest                       & 0.00 $\pm$ 0.00             & 120.00 $\pm$ 0.00    & \textbf{180.00 $\pm$ 0.00}           & \textbf{180.00 $\pm$ 0.00}                   & -          & -             & -          \\
space\_invaders                & \textbf{398.00 $\pm$ 16.65} & 278.00 $\pm$ 11.71   & 285.00 $\pm$ 0.00           & 273.17 $\pm$ 11.93            & -30.20\%                       & -28.40\%                          & -31.40\%                       \\
venture                        & 0.00 $\pm$ 0.00             & 0.00 $\pm$ 0.00      & 0.00 $\pm$ 0.00             & 0.00 $\pm$ 0.00               & 0\%                            & 0\%                               & 0\%                            \\
\bottomrule
average                        & -                         & -                  & -                         & -                           & 211.7\%                        & 212.9\%                           & \textbf{343.6\%}   \\
\bottomrule
\end{tabular}
}
\label{tab:bco_300}
\end{table*}

\begin{table*}[]
\caption{Game scores obtained when using 30 minutes of human demonstration data to train the T-REX agents. Results are presented as mean$\pm$standard error of the mean (N=30). The improvement columns show the relative improvement over the T-REX baseline. ``-" indicates that the baseline score is zero hence the relative improvement is not calculated and is not counted in the average. Average results over all 20 games are presented in Table \ref{tab:trex_improv}.}
\begin{tabular}{c|ccc}
\toprule
        & T-REX  & T-REX +CGL                     & Improv-CGL \\
        \midrule
alien                          & 727.00 $\pm$ 52.24          & \textbf{800.33 $\pm$ 66.44}     & 10.1\%                         \\
asterix & 5023.33 $\pm$ 431.17        & \textbf{66445.00 $\pm$ 7444.18} & 1222.7\%                       \\
bank\_heist                    & 0.00 $\pm$ 0.00             & \textbf{19.33 $\pm$ 3.86}       & -          \\
berzerk                        & 273.33 $\pm$ 10.20          & \textbf{584.00 $\pm$ 22.18}     & 113.7\%                        \\
breakout                       & 46.33 $\pm$ 1.75            & \textbf{386.77 $\pm$ 25.42}     & 734.8\%                        \\
centipede                      & 8369.30 $\pm$ 971.79        & \textbf{20762.47 $\pm$ 2120.94} & 148.1\%                        \\
demon\_attack                  & 3846.00 $\pm$ 513.18  & \textbf{6767.33 $\pm$ 783.84}                 & 75.9\%                       \\
enduro                         & 0.00 $\pm$ 0.00             & 0.00 $\pm$ 0.00                 & 0.0\%                          \\
freeway                        & 0.00 $\pm$ 0.00             & \textbf{0.07 $\pm$ 0.05}        & -          \\
frostbite                      & 1.00 $\pm$ 0.55             & \textbf{36.67 $\pm$ 1.28}       & 3567.0\%                       \\
hero                           & 0.00 $\pm$ 0.00             & 0.00 $\pm$ 0.00                 & 0.0\%                          \\
montezuma\_revenge             & 0.00 $\pm$ 0.00             & 0.00 $\pm$ 0.00                 & 0.0\%                          \\
ms\_pacman                     & \textbf{967.00 $\pm$ 84.22} & 577.33 $\pm$ 61.54              & -40.3\%                        \\
name\_this\_game               & 2262.00 $\pm$ 104.89        & \textbf{4081.00 $\pm$ 175.34}   & 80.4\%                         \\
phoenix                        & 303.67 $\pm$ 28.24          & \textbf{502.67 $\pm$ 39.24}     & 65.5\%                         \\
riverraid                      & 1748.00 $\pm$ 31.49         & \textbf{5201.67 $\pm$ 203.35}   & 197.6\%                        \\
road\_runner                   & 0.00 $\pm$ 0.00             & \textbf{2660.00 $\pm$ 359.77}   & -          \\
seaquest                       & 0.00 $\pm$ 0.00             & \textbf{759.33 $\pm$ 12.56}     & -          \\
space\_invaders                & 607.00 $\pm$ 44.32          & \textbf{923.50 $\pm$ 59.82}     & 52.1\%                         \\
venture                        & 0.00 $\pm$ 0.00             & 0.00 $\pm$ 0.00                 & 0.0\%                          \\
\bottomrule
average                        & -                         &     -                          & 390.4\%\\
\bottomrule
\end{tabular}
\label{tab:trex_30}
\end{table*}

\begin{table*}[]
\caption{Game scores obtained when using 300 minutes human demonstration data to train the T-REX agents. Results are presented as mean$\pm$standard error of the mean (N=30). The improvement columns show the relative improvement over the T-REX baseline. ``-" indicates that the baseline score is zero hence the relative improvement is not calculated and is not counted in the average. Average results over all 20 games are presented in Table \ref{tab:trex_improv}.}
\begin{tabular}{c|ccc}
\toprule
                         & T-REX      & T-REX +CGL                     & Improv-CGL \\
\midrule
alien                          & 359.67 $\pm$ 8.48               & \textbf{1007.33 $\pm$ 48.94}    & 180.10\%                       \\
asterix & 15231.67 $\pm$ 2401.26          & \textbf{17073.33 $\pm$ 2253.10} & 12.10\%                        \\
bank\_heist                    & 2.33 $\pm$ 0.77                 & \textbf{7.00 $\pm$ 1.17}        & 200.40\%                       \\
berzerk                        & 411.67 $\pm$ 40.87              & \textbf{596.67 $\pm$ 32.52}     & 44.90\%                        \\
breakout                       & 53.33 $\pm$ 1.30                & \textbf{438.40 $\pm$ 17.59}     & 722.10\%                       \\
centipede                      & \textbf{16363.07 $\pm$ 1993.35} & 13532.70 $\pm$ 1550.44          & -17.30\%                       \\
demon\_attack                  & 463.17 $\pm$ 138.3              & \textbf{17589.00 $\pm$ 1727.02} & 3697.50\%                      \\
enduro                         & \textbf{0.90 $\pm$ 0.57}        & 0.67 $\pm$ 0.16                 & -25.60\%                       \\
freeway                        & 0.00 $\pm$ 0.00                 & \textbf{0.07 $\pm$ 0.05}        &-          \\
frostbite                      & 22.67 $\pm$ 1.88                & \textbf{208.00 $\pm$ 6.28}      & 817.50\%                       \\
hero                           & 0.00 $\pm$ 0.00                 & \textbf{2.50 $\pm$ 2.46}        &-         \\
montezuma\_revenge             & 0.00 $\pm$ 0.00                 & 0.00 $\pm$ 0.00                 & 0\%        \\
ms\_pacman                     & 314.00 $\pm$ 22.41              & \textbf{527.33 $\pm$ 38.74}     & 67.90\%                        \\
name\_this\_game               & 3331.67 $\pm$ 185.85            & \textbf{4010.67 $\pm$ 147.50}   & 20.40\%                        \\
phoenix                        & \textbf{4322.33 $\pm$ 363.98}   & 2123.67 $\pm$ 164.78            & -50.90\%                       \\
riverraid                      & 5812.67 $\pm$ 233.48            & \textbf{7370.00 $\pm$ 262.18}   & 26.80\%                        \\
road\_runner                   & 0.00 $\pm$ 0.00                 & \textbf{1286.67 $\pm$ 111.73}   &-         \\
seaquest                       & 0.00 $\pm$ 0.00                 & \textbf{729.33 $\pm$ 16.32}     &-          \\
space\_invaders                & 410.50 $\pm$ 46.54              & \textbf{1563.67 $\pm$ 144.76}   & 280.90\%                       \\
venture                        & 0.00 $\pm$ 0.00                 & 0.00 $\pm$ 0.00                 & 0\% \\             
\bottomrule
average                        & -                         &     -                          & 373.6\%\\
\bottomrule
\end{tabular}
\label{tab:trex_300}
\end{table*}

\begin{table*}[]
\caption{Confounding study results. Game scores are obtained when using 15 minutes of human demonstration data to train the agents. Results are presented as mean$\pm$standard error of the mean (N=30). The change columns show the relative change over the non-confounded baselines. On average CGL agents suffer less when trained with confounded data, and still perform better than behavioral cloning (BC) agents. These results are discussed in Sec. \ref{sec:causal}.}
\resizebox{\textwidth}{!}{
\begin{tabular}{c|cccccccc}
\toprule
& BC             & CGL            & Improv-CGL & BC-confounded & CGL-confounded & Improv-CGL-confounded & Change-BC & Change-CGL\\
\midrule
alien              & 1575$\pm$176.8    & 2044.7$\pm$242.1  & 29.8\%                         & 73$\pm$9.3       & 439.3$\pm$63.4    & 501.8\%                        & -95.4\%                                            & -78.5\%                                             \\
asterix            & 285$\pm$28.2      & 426.7$\pm$27.8    & 49.7\%                         & 243.3$\pm$23.1   & 363.3$\pm$30      & 49.3\%                         & -14.6\%                                            & -14.9\%                                             \\
bank\_heist        & 86.3$\pm$9.0      & 143$\pm$14.8      & 65.7\%                         & 22.3$\pm$3.2     & 15.3$\pm$2.8      & -31.4\%                        & -74.2\%                                            & -89.3\%                                             \\
berzerk            & 330.7$\pm$22.0    & 366.7$\pm$19      & 10.9\%                         & 101.7$\pm$11.2   & 322$\pm$21.5      & 216.6\%                        & -69.2\%                                            & -12.2\%                                             \\
breakout           & 2.2$\pm$0.3       & 3.7$\pm$0.4       & 68.2\%                         & 0$\pm$0          & 0.4$\pm$0.1       &-        & -100.0\%                                           & -89.2\%                                             \\
centipede          & 4378.8$\pm$442.6  & 6075.9$\pm$845.1  & 38.8\%                         & 5320.8$\pm$705.5 & 5808.5$\pm$640.5  & 9.2\%                          & 21.5\%                                             & -4.4\%                                              \\
demon\_attack      & 112.2$\pm$13.9    & 205.2$\pm$41.9    & 82.9\%                         & 120.5$\pm$10     & 200.7$\pm$32.1    & 66.6\%                         & 7.4\%                                              & -2.2\%                                              \\
enduro             & 11.7$\pm$2.0      & 4.2$\pm$1.4       & -64.1\%                        & 3.3$\pm$0.7      & 8.3$\pm$1.5       & 151.5\%                        & -71.8\%                                            & 97.6\%                                              \\
freeway            & 29.4$\pm$0.2      & 30$\pm$0.3        & 2.0\%                          & 23.9$\pm$0.2     & 26.5$\pm$0.2      & 10.9\%                         & -18.7\%                                            & -11.7\%                                             \\
frostbite          & 1628.3$\pm$246.4  & 2973$\pm$279      & 82.6\%                         & 189.3$\pm$3.7    & 710$\pm$99.3      & 275.1\%                        & -88.4\%                                            & -76.1\%                                             \\
hero               & 13255.3$\pm$845.1 & 19023.2$\pm$679.7 & 43.5\%                         & 109.7$\pm$97.8   & 10038.3$\pm$647.6 & 9050.7\%                       & -99.2\%                                            & -47.2\%                                             \\
montezuma\_revenge & 100$\pm$31.6      & 1200$\pm$159.2    & 1100.0\%                       & 0$\pm$0          & 0$\pm$0           & 0.0\%                          & -100.0\%                                           & -100.0\%                                            \\
ms\_pacman         & 843.3$\pm$62.8    & 1348.3$\pm$206.9  & 59.9\%                         & 281$\pm$65.4     & 397$\pm$31        & 41.3\%                         & -66.7\%                                            & -70.6\%                                             \\
name\_this\_game   & 1917.3$\pm$130.2  & 2646.3$\pm$156.3  & 38.0\%                         & 2204.7$\pm$196.1 & 3179.3$\pm$173.4  & 44.2\%                         & 15.0\%                                             & 20.1\%                                              \\
phoenix            & 1060$\pm$172.4    & 2193.7$\pm$200.5  & 107.0\%                        & 550.3$\pm$76.8   & 1657$\pm$263.9    & 201.1\%                        & -48.1\%                                            & -24.5\%                                             \\
riverraid          & 2771.7$\pm$141.8  & 2965.3$\pm$184.8  & 7.0\%                          & 2457$\pm$113.8   & 2105$\pm$67.1     & -14.3\%                        & -11.4\%                                            & -29.0\%                                             \\
road\_runner       & 7840$\pm$553.3    & 12723.3$\pm$376.7 & 62.3\%                         & 6290$\pm$364.1   & 10023.3$\pm$396   & 59.4\%                         & -19.8\%                                            & -21.2\%                                             \\
seaquest           & 194$\pm$11.4      & 216$\pm$11.2      & 11.3\%                         & 182.7$\pm$10.3   & 182$\pm$10.1      & -0.4\%                         & -5.8\%                                             & -15.7\%                                             \\
space\_invaders    & 275$\pm$26.2      & 314$\pm$26        & 14.2\%                         & 219.3$\pm$21.9   & 265.5$\pm$26.9    & 21.1\%                         & -20.3\%                                            & -15.4\%                                             \\
venture            & 196.7$\pm$26.5    & 376.7$\pm$16.1    & 91.5\%                         & 6.7$\pm$6.6      & 20$\pm$11         & 198.5\%                        & -96.6\%                                            & -94.7\%                                             \\
\bottomrule
average            & -              & -              & \textbf{95.1\%}                         & -             & -              & \textbf{571.1\%}                        & -47.8\%                                            & \textbf{-34.0\%} \\                                           
\bottomrule
\end{tabular}}
\label{tab:bc_confounded}
\end{table*}

\end{document}